\documentclass[runningheads]{llncs}
\usepackage{graphicx}
\usepackage{url}
\usepackage{graphicx}
\usepackage{multirow}
\usepackage{amsmath}
\usepackage{amssymb}

\let\llncssubparagraph\subparagraph
\let\subparagraph\paragraph
\usepackage[compact]{titlesec}
\let\subparagraph\llncssubparagraph

\titlespacing{\section}{0pt}{10pt plus 1pt minus 1pt}{5pt plus 1pt}
\titlespacing{\subsection}{0pt}{10pt plus 1pt minus 1pt}{5pt plus 1pt}
\usepackage[colorlinks,
            linkcolor=blue,
            anchorcolor=blue,
            citecolor=blue]{hyperref}
\makeatletter
\def\thanks#1{\protected@xdef\@thanks{\@thanks
        \protect\footnotetext{#1}}}
\makeatother
\begin{document}
\title{Representation and Correlation Enhanced Encoder-Decoder Framework for Scene Text Recognition}
% \thanks{*\href{https://github.com/Mona9955/RCEED-ICDAR2021}{https://github.com/Mona9955/RCEED-ICDAR2021}}}
\author{Mengmeng Cui\inst{1}, Wei Wang\inst{1}, Jinjin Zhang\inst{1}, Liang Wang\inst{1,2}}
\institute{Institute of Automation, Chinese Academy of Sciences (CASIA)  \and  School of Artificial Intelligence, University of Chinese Academy of Sciences (UCAS) \\ \email{\{mengmeng.cui, jinjin.zhang\}@cripac.ia.ac.cn} \\ \email{\{wangwei, wangliang\}@nlpr.ia.ac.cn}}
\titlerunning{RCEED}
\authorrunning{}
\maketitle              % typeset the header of the contribution
\begin{abstract}
Attention-based encoder-decoder framework is widely used in the scene text recognition task. However, for the current state-of-the-art(SOTA) methods, there is room for improvement in terms of the efficient usage of local visual and global context information of the input text image, as well as the robust correlation between the scene processing module(encoder) and the text processing module(decoder). In this paper, we propose a Representation and Correlation Enhanced Encoder-Decoder Framework(RCEED) to address these deficiencies and break performance bottleneck. In the encoder module, local visual feature, global context feature, and position information are aligned and fused to generate a small-size comprehensive feature map. In the decoder module, two methods are utilized to enhance the correlation between scene and text feature space. 1) The decoder initialization is guided by the holistic feature and global glimpse vector exported from the encoder. 2) The feature enriched glimpse vector produced by the Multi-Head General Attention is used to assist the RNN iteration and the character prediction at each time step. Meanwhile, we also design a Layernorm-Dropout LSTM cell to improve model's generalization towards changeable texts. Extensive experiments on the benchmarks demonstrate the advantageous performance of RCEED in scene text recognition tasks, especially the irregular ones. 
\end{abstract}
\section{Introduction}
Scene Text Recognition(STR) refers to the text recognition of natural scene images captured by camera. Compared with traditional Optical Character Recognition(OCR) systems dedicated to high-quality document images, STR techniques are developed for the outdoor images and applied in a wider range of fields, such as street view positioning, image advertisement filtering, bill recognition, et al. Due to the randomness in the process of capturing text images in natural scenes, STR has many challenges in practical applications, including uneven lighting and focusing caused image quality degradation, complex image background, occluded and incomplete characters. Moreover, the characters themselves also have diverse font types, font sizes and colors; the text distribution is irregular, many of which are perspective, distorted or oriented. Therefore, as a complex problem, STR has been extensively studied in industry and academia.

STR is divided into regular and irregular text recognition tasks according to the text distribution. Modern technical solutions mainly include Connectionist Temporal Classification(CTC) based methods\cite{shi2016end}, attention-based encoder-decoder framework\cite{shi2018aster,yang2020holistic} and the combination of both\cite{litman2020scatter,zuo2019natural}. These methods only need word-level annotations and robust to complicated scene text images. CTC-based methods solve the problem of misalignment between the input image and the target outputs but can not leverage the contextual dependency between characters, thus it is mainly used for horizontal text recognition. Comparatively, attention mechanism is a good way to strengthen the relevance between visual and semantic features and improve the interpretability of the model, making it a suitable choice for irregular STR scenarios. The major architectures of the attention-based encoder-decoder framework include the sequence-to-sequence models which adopt 1D attention mechanism\cite{shi2018aster} or 2D attention mechanism\cite{li2019show} in the decoder, and the transformer-based models\cite{yang2020holistic,lu2019master,sheng2019nrtr}.

However, there are still shortcomings for the existing attention-based encoder-decoder framework. 1D attention methods generally use RNN layer(s) to model contextual dependencies but lose apparent information of the text\cite{shi2018aster}. Although 2D attention is able to handle the irregular spatial distribution of the text, its performance is greatly restricted by the size of the encoded feature map\cite{li2019show}. Since the inter-character dependence in STR is weaker than the inter-word dependence in machine translation, the self-attention design in transformer-based models which targets at building long-range dependencies may not achieve the expected performance\cite{yang2020holistic}, but increases the parameters due to its multiple fully-connected layers and the multi-layer stacking structure of decoder. The common reasons for these problems lie in the information loss of the global and local feature, as well as the weak relevance between the encoder working on the visual space and the decoder working on the language space. 

Therefore, we propose the $\mathbf{R}$epresentation and $\mathbf{C}$orrelation $\mathbf{E}$nhanced $\mathbf{E}$ncoder-$\mathbf{D}$ecoder Framework(RCEED). The encoder generates a comprehensive representation of local visual feature and global context feature. The decoder utilizes the Multi-Head General Attention mechanism to capture an enriched glimpse of the encoded feature. The initialization manner and the efficient workflow of the decoder increase the correlation between the visual feature and the decoded characters.  
Our main contributions are summarized as follows:
\\ 1. \ In the encoder module, a representation enhanced feature map is obtained by combining the visual, context and position information. The encoded feature map has a small size corresponding to the spatial distribution of characters. 
\\ 2. \ In the decoder module, a holistic feature and the a global glimpse vector are introduced from the encoder to guide the initialization of the decoder. The intuitive workflow enables the glimpse vector to participate in the update of the decoded hidden state and the character prediction at the same time. These integrated designs make the model achieve SOTA performance in public benchmarks, especially the irregular ones.
\\ 3. \ We devise a Mulit-Head General Attention mechanism to capture the main information and the supplementary information of the encoded feature with fewer operations and parameters.
\\ 4. \  We specially design the LD-LSTM cell as basic block to form the RNN layers of the encoder and the decoder. The LD-LSTM can balance independence and relevance between characters and improve model's generalization for irregular texts, which is very important for the STR applications. 

\section{Related Work}
Early text image recognition is oriented to the document recognition scenario which has a clear picture and a fixed pattern. People use binarization method\cite{casey1996survey} and sliding window method\cite{wang2011end} for individual character detection, and then integrate the characters into words by dynamic programming. These methods are vulnerable to the background noise, and unable to use the global context information. Later works tend to treat the text image recognition as a sequence recognition problem. These methods are more capable of the complex STR tasks and mainly divided into two categories, the CTC-based methods and the attention based methods. CRNN\cite{shi2016end} utilizes the CNN and RNN to generate feature sequence from the visual information, and CTC to align the characters predicted by the RNN decoder. Attention based encoder-decoder models like RARE\cite{shi2016robust} and $R^2AM$\cite{lee2016recursive} are developed to introduce the attention-mechanism from machine translation\cite{bahdanau2014neural} to solve the image-based sequence recognition problem. Focusing attention network\cite{cheng2017focusing} is raised to fix the attention drifting caused by complex scenes or low image quality. With similar targets, Bai et al. introduce the edit probability\cite{bai2018edit} method to alleviate character missing or superfluous in text recognition. In addition, model as a combination of CTC and attention based methods\cite{zuo2019natural} also performs well on regular scene text datasets.

In recent years, many approaches have been proposed regarding the more challenging irregular scene text recognition task. The first type is the rectification based methods. ASTER\cite{shi2018aster} combines the Thin-Plate Spline (TPS) method\cite{bookstein1993thin} and the Spatial Transformer Network(STN)\cite{jaderberg2015spatial} to form the rectification network. The line-fitting transformation method is proposed in ESIR\cite{zhan2019esir} which employs iterative rectification to improve the performance. MORAN\cite{luo2019moran} proposes a pixel transform method to make a smooth conversion to the text images. The other type is the character level methods. Models like Char-Net\cite{liu2018char} and Mask TextSpotter\cite{liao2019mask} detect and rectify the individual characters, which requires additional character-level supervision. The last type is the attention based encoder-decoder frameworks. 2D feature map is used for both the sequence-to-sequence models\cite{li2019show,yue2020robustscanner} and the transformer-based models\cite{yang2020holistic,lu2019master}. The expanded focus range contributes to the recognition of characters with arbitrary shape and position. RobustScanner\cite{yue2020robustscanner} introduces a positional enhancement branch to the 2D attentional encoder-decoder structure proposed in SAR\cite{li2019show}, leveraging the positional information to make prediction during decoding process. MASTER\cite{lu2019master} employs the non-local network as the encoder of the transformer-based structure to capture longer contextual dependencies.

\section{Model Architecture}
\begin{figure}
\vspace{-0.3cm}
\setlength{\belowcaptionskip}{-0.5cm}
\includegraphics[width=\textwidth]{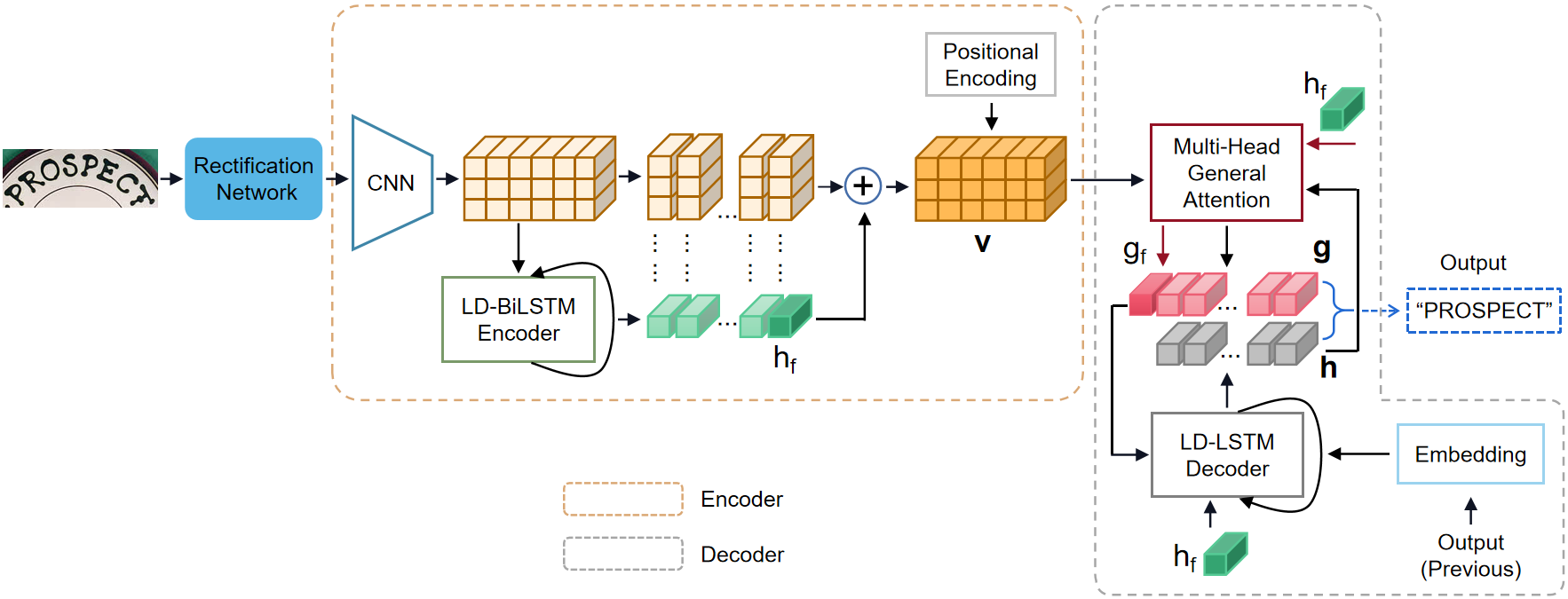}
\caption{Architecture of the RCEED. The input image is transformed by the Rectification Network before fed into the encoder. The encoder combines the visual feature from CNN, the context feature from the LD-BiLSTM layer and the position information to generate the comprehensive feature map($\mathbf{v}$). The holistic feature($h_{f}$) and the global glimpse vector($g_{f}$) from encoder are used to guide decoder initialization. At each decoding step t, the glimpse vector $g_{t}( \in \mathbf{g})$ is calculated by MHGAT based on the current hidden state $h_{t}( \in \mathbf{h})$, and exported to the next LD-LSTM iteration to generate $h_{t+1}$. Both $g_{t}$ and $h_{t}$ are used to make the current prediction.} 
\label{Fig.1}
\end{figure}

As presented in Figure \ref{Fig.1}, there are three important components of RCEED: the Rectification Network which redistributes the characters, the Representation Enhanced Encoder which combines the local visual feature and global context feature, the Multi-Head General Attention Decoder which increases the correlation between the visual space and the language space. Two basic composition methods are utilized in the encoder and decoder, including the LD-LSTM cell which improves model's generalization and the Multi-Head General Attention mechanism which makes effective use of main and supplementary information.
\subsection{Rectification Network}
Thin-Plat Spline\cite{bookstein1993thin} is a 2D interpolation method which makes conversion with minimal bending energy based on a set of corresponding control points of two pictures. Similar to the STN in \cite{shi2018aster}, our rectification network utilizes a lightweight CNN to generate the control points of the source image, associating them with the pre-defined control points of the target image through TPS to accomplish the rectification process. We also find that the rectification network is able to adjust the width and spacing of characters adaptively, which reduces the divergence of input text images and improves the alignment between receptive fields and characters.

\subsection{Basic Composition Methods for Encoder-Decoder Framework}
In order to facilitate understanding, before the illustration of the Encoder and Decoder, we introduce two methods which are important components of them. 
\subsubsection{Layernorm-Dropout LSTM Cell(LD-LSTM)}
The Long Short-Term Memory(LSTM) \cite{hochreiter1997long} is widely used in the machine translation models\cite{bahdanau2014neural}. Considering the difference between the text recognition task and the language processing task which has strong semantic dependencies between tokens, we specially design the Layernorm-Dropout LSTM Cell.

Firstly, we add layernorm\cite{ba2016layer} to the current input and the previous hidden state of the LSTM cell to speed up the convergence of training process and regularize the network. Secondly, on account of the relatively weak dependencies between characters in the text image, we introduce the dropout function\cite{hinton2012improving} to reduce feature co-adaptation and improve the presentation of important information. Different from the conventional way of applying dropout in the feed-forward connections between RNN layers\cite{zaremba2014recurrent}, we design a per-step dropout method in the recurrent connections of RNN cells. Both the hidden state and the cell state are sampled by the dropout masks with probability p to balance the relevance and independence between characters, and improve the generalization at the same time. 

The layernorm operation is given by:
\begin{equation}
    LN(\mathbf{x}; \alpha ,\beta )=\frac{(\mathbf{x}-\mu )}{\sigma }\odot \alpha +\beta 
\end{equation}
\begin{equation}
    \mu =\frac{1}{D}\sum_{i=1}^{D}x_{i}, \quad\sigma =\sqrt{\frac{1}{D}\sum_{i=1}^{D}(x_{i}-\mu )^{2}}
\end{equation}

Where $x_{i}$ is the $i_{th}$ element of $\mathbf{x}$ with length D. $\alpha$ and $\beta$ are defined as gain and bias parameters. In this paper, we set $\beta$ as zero. Then the LD-LSTM cell is defined as:
\begin{equation}
\label{eq_3}
\begin{pmatrix}
f_{t}\\ 
i_{t}\\ 
o_{t}\\ 
\hat{c}_{t}
\end{pmatrix}= LN(W_{x}x_{t}; \alpha _{1}, \beta _{1}) + LN(W_{h}h_{t-1}; \alpha _{2}, \beta _{2})
\end{equation}
\begin{equation}
c_{t} = Dropout\left (sigm(f_{t})\odot c_{t-1}+sigm(i_{t})\odot tanh(\hat{c}_{t}), \ p \right)
\end{equation}
\begin{equation}
h_{t} = Dropout\left (sigm(o_{t})\odot tanh(c_{t}), \ p \right )
\end{equation}

Where $W_x$ and $W_h$ are weight matrixes of the input $x_t$ and the hidden state $h_{t-1}$. $\odot$ denotes the element-wise product operation. $p$ is the probability of each element of the vector being zero.

\subsubsection{Multi-Head General Attention(MHGAT)}
The commonly used attention functions include general attention, additive attention, dot product attention et al. Unlike most sequence-to-sequence text recognition models which utilize additive attention mechanism to build connection between encoder and decoder\cite{shi2018aster,litman2020scatter,li2019show}, we use general function\cite{luong2015effective} to reduce the computational complexity. Compared with additive attention\cite{bahdanau2014neural} which needs two-step operations to obtain the attention weights(add first and then multiply by the transform matrix), general attention only needs one-step matrix multiplication operation(Equation (\ref{eq_6})). Given $\mathbf{v^{'}}$=[$v_{1}^{'}, v_{2}^{'},...,v_{N}^{'}$] as a splitted part of the flattened comprehensive feature map $\mathbf{v}$ with length N, the formulation of the general attention mechanism for $\mathbf{v^{'}}$ can be expressed as follows:
\begin{equation}
\label{eq_6}
   score(h_{t}^{'}, \mathbf{v^{'}}) = \frac{\mathbf{v}W_{a}^{'}h_{t}^{'}}{d_{v^{'}}^{2}} 
\end{equation}
\begin{equation}
\mathbf{a_{t}^{'}}=softmax(score(h_{t}^{'}, \mathbf{v^{'}})) \in \mathbb{R}^{N}
\end{equation}
\begin{equation}
GeneralAttention(h_{t}^{'}, \mathbf{v^{'}}) = \sum_{i=1}^{N}a_{t,i}^{'}v_{i}^{'}
\end{equation}

Where $a_{t,i}^{'}$ is the $i_{th}$ element of the attention weights $\mathbf{a_{t}^{'}}$. $W_{a}^{'}\in \mathbb{R}^{d_{v}\times d_{v^{'}}}$ is a parameter matrix with $d_{v}$ and $d_{v^{'}}$ representing the dimensions of $\mathbf{v}$ and $\mathbf{v^{'}}$. In order to build a harder attention module to suppress the background noise, we set the a relatively large scale factor, which is  $d_{v^{'}}^{2}$ under 8 parallel heads.

Due to the diversity of character size and distribution, there is misalignment between the feature map obtained after CNN downsampling and the character visual information. Therefore, we adopt the multi-head attention(shown in Figure \ref{Fig.4}(a)) to increase the attention flexibility and reduce information loss. Given the hidden state $h_{t}$ at time step t as a query, the Multi-Head General Attention for $\mathbf{v}$ is generated by:
\begin{equation}
    g_{t}=MultiHead(h_{t}, \mathbf{v}) = Concat(head_{1},...,head_{m}) 
\end{equation}
\begin{equation}
where \ head_{j} = GeneralAttention(h_{t}W_{h,j}, \mathbf{v_{j}}), \ \mathbf{v_{j}} \in Split(\mathbf{v}, m) \notag
\end{equation}

\begin{figure}
\centering
\includegraphics[width=\textwidth]{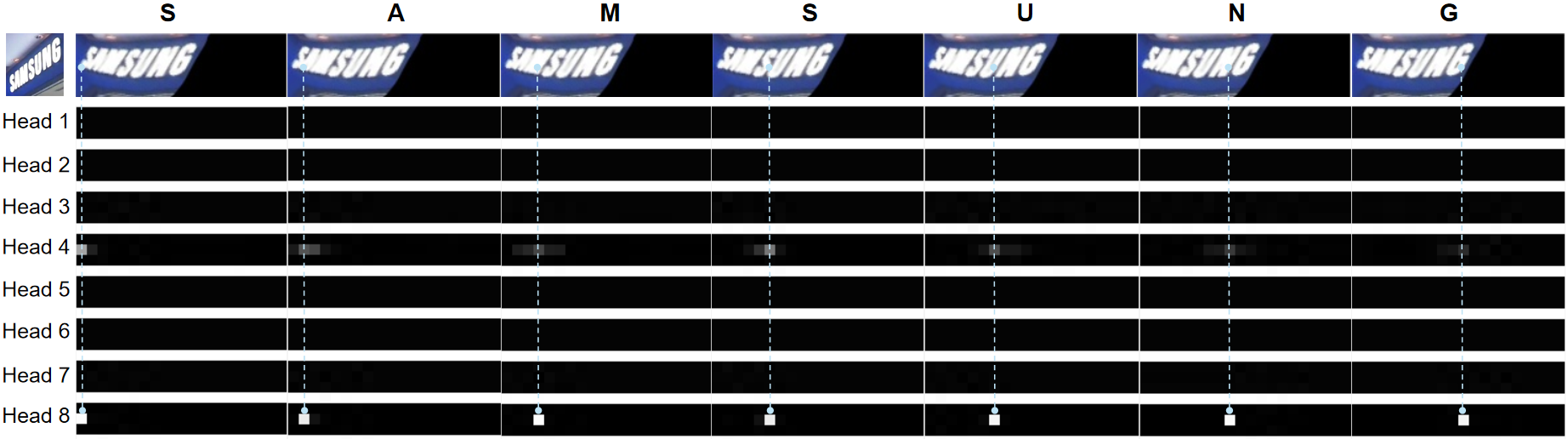}
\caption{Visualization of the 8-heads attention map at each decoding step. Head 8 is in charge of the main information and completely aligned with the decoded characters. Head 4 pays attention to the supplementary information around the target area.} 
\label{Fig.2}
\end{figure}

The parameters are $W_{h,j} \in  \mathbb{R}^{d\times \frac{d_{v}}{m}}$, where $d$ denotes the dimension of $h_{t}$, m is the number of parallel heads. $\mathbf{v_{j}}$ is obtained by splitting m times along $d_{v}$. $g_{t}$ as a concatenation of the attention heads refers to the "glimpse" of the encoded feature where the current step attends to. Figure \ref{Fig.2} visualizes the 8-heads attention maps at each decoding step. Head 8 and head 4 are respectively in charge of the main and the supplementary attention tasks, and accurately aligned with the target characters.

\subsection{Representation Enhanced Encoder}
The encoder of RCEED consists of two parts: the Resnet[18] based CNN backbone(presented in Table \ref{tab1}) which extracts local visual feature, and the single-layer LD-BiLSTM which outputs global context feature. Limited by the receptive field, the CNN encoder can not well represent the contextual information. Therefore, BiLSTM\cite{hochreiter1997long} layer with the LD-LSTM Cell as block(abbreviated as LD-BiLSTM) is utilized to generate the context feature sequence, which is based on the intermediate feature sequence produced by average pooling on the visual feature column vectors. The final hidden state is taken as the holistic feature.

\begin{figure}
\centering
\includegraphics[width=0.6\textwidth]{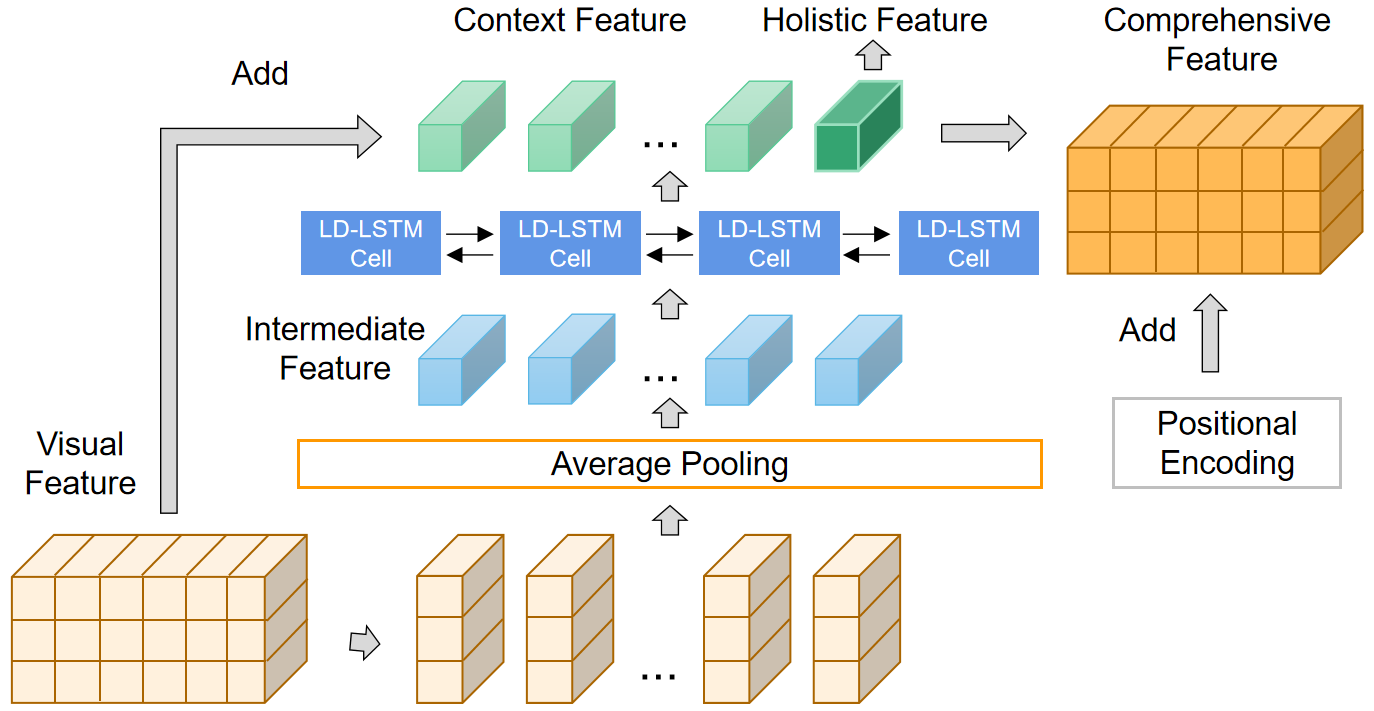}
\caption{Structure of the Representation Enhanced Encoder. The intermediate feature sequence is obtained by applying average pooling on the visual feature column vectors. After the column-aligned addition between the visual feature and the context feature, as well as the positional encoding operation, a comprehensive feature map is generated.} 
\label{Fig.3}
\end{figure}

However, due to the abstractness of the LD-BiLSTM operation, the complex shape and spatial information of the text is lost. Thus, as shown in Figure \ref{Fig.3}, we add the context feature sequence and the corresponding visual feature column vectors along the row axis, getting the comprehensive feature representation. 
\begin{table}[]
\caption{The configuration of the ResNet based CNN feature extractor. The stride and padding for the convolutional layers are all set to 1. The 'k' and 's' in the max-pooling layers refer to kernel size and stride. Height and width of the output of each layer are presented in the last column.}
\label{tab1}
\centering
\begin{tabular}{c|c|c}
\hline
\textbf{Layer} & \textbf{\quad Configuration \quad} & \textbf{\quad Output \quad} \\ \hline
Conv           & $3\times 3, \ 64$                      & $48\times 160$               \\
Conv           & $3\times 3, \ 128$                      & $48\times 160$               \\
Max-pooling    & $k:2\times 2, \ s:2\times 2$                      & $24\times 80$               \\
Residual block \quad \quad & $\begin{bmatrix}
\ 3\times 3, \ 256 \ \\ 
\ 3\times 3, \ 256 \ 
\end{bmatrix} \ \times  \ 1$                      & $24\times 80$               \\
Conv           & $3\times 3, \ 256$                       & $24\times 80$               \\
Max-pooling    & $k:2\times 2, \ s:2\times 2$                      & $12\times 40$               \\
Residual block & $\begin{bmatrix}
\ 3\times 3, \ 256 \ \\ 
\ 3\times 3, \ 256 \ 
\end{bmatrix} \ \times  \ 2$                      &  $12\times 40$              \\
Conv           & $3\times 3, \ 256$                      & $12\times 40$               \\
Max-pooling    & $k:2\times 2, \ s:2\times 2$                      & $6\times 20$               \\
Residual block & $\begin{bmatrix}
\ 3\times 3, \ 512 \ \\ 
\ 3\times 3, \ 512 \ 
\end{bmatrix} \ \times  \ 5$                      & $6\times 20$               \\
Conv           & $3\times 3, \ 512$                      & $6\times 20$               \\
Max-pooling    & $k:2\times 2, \ s:[2,1]\times [2,1]$                      & $3\times 20$               \\
Residual block & $\begin{bmatrix}
\ 3\times 3, \ 512 \ \\ 
\ 3\times 3, \ 512 \ 
\end{bmatrix} \ \times  \ 3$                      & $3\times 20$               \\
Conv           & $3\times 3, \ 512$                      & $3\times 20$               \\ \hline
\end{tabular}
\end{table}

Since the character position distribution along the vertical direction is mainly divided into three parts: upper, middle, and lower. Meanwhile, for the sake of reducing computational consumption, we set the feature map size as 3$\times$20, where 3 corresponds to the 3 types vertical position. Compared with \cite{li2019show,lu2019master,yue2020robustscanner}, our feature map size is only a quarter of theirs. Similar to \cite{vaswani2017attention}, we consider the influence of position information and use sinusoid function to encode absolute as well as relative position information to the comprehensive feature map.

\subsection{Recurrent Multi-Head General Attention Decoder}
\begin{figure}
\vspace{-0.2cm}
\setlength{\belowcaptionskip}{-0.6cm} 
\centering
\includegraphics[width=\textwidth]{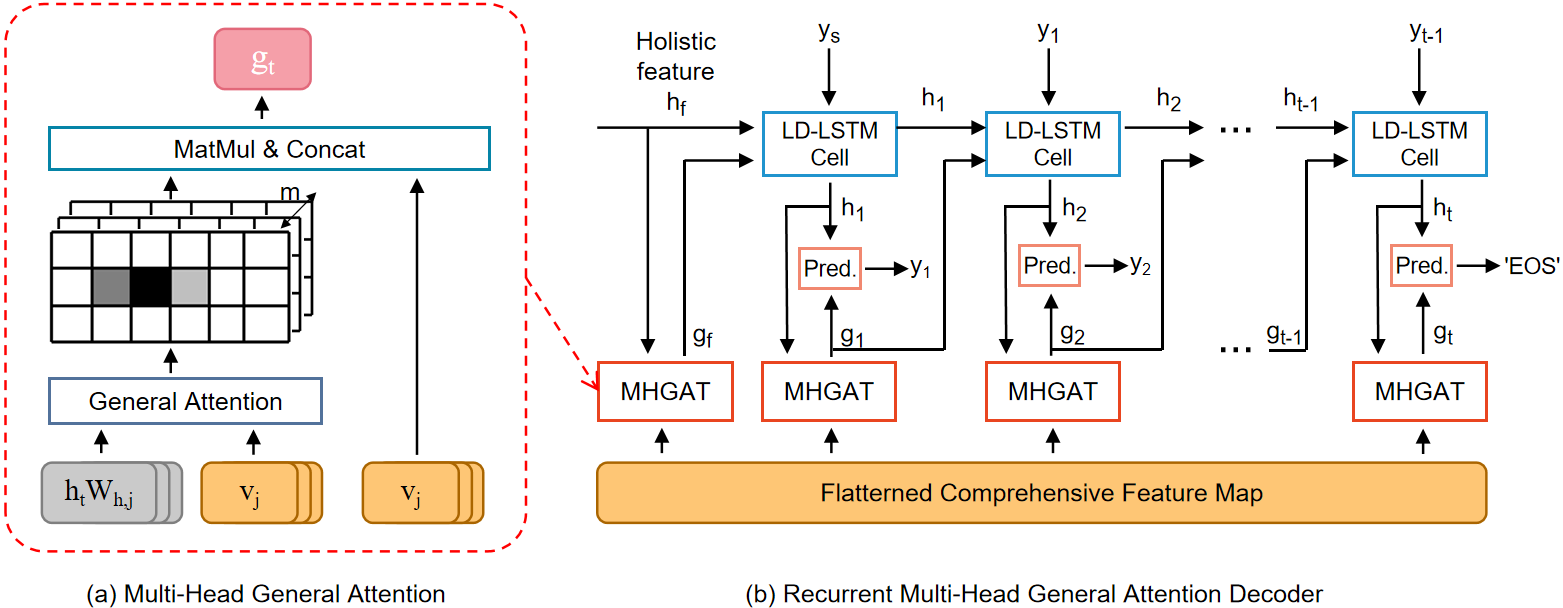}
\caption{(a) Structure of the Multi-Head General Attention.  (b) Structure of the  Recurrent Multi-Head General Attention Decoder. When decoding starts, the holistic feature $h_{f}$ and global glimpse vector $g_{f}$ from encoder are input to the LD-LSTM cell together with the start token $y_{s}$ to get the hidden state $h_{1}$. $h_{1}$ and its related glimpse vector $g_{1}$ are both used to make the first prediction. Then decoding iteration continues based on the previous outputs until the 'EOS' token is predicted.} 
\label{Fig.4}
\end{figure}

As shown in Figure \ref{Fig.4}(b), the decoder is built on a sequence-to-sequence architecture with the LD-LSTM cell and the Multi-Head General Attention as basic blocks. Before the decoding starts, given the flattened comprehensive feature map $\mathbf{v}$ and the holistic feature $h_{f}$ from the encoder as inputs, a global glimpse vector $g_{f}$ related to $h_{f}$ is derived from MHGAT. $h_{f}$ and $g_{f}$ represent the global information of the encoded feature map and serve as the guidance for the initialization of decoder, which helps to focus on the explicit part at the first decoding step and improve the accuracy. Then, the LD-LSTM operation is performed fed with $g_{f}$, $h_{f}$, and a start token $y_{s}$. Note that the LD-LSTM is slightly changed in the decoder, the word embedding of the start token and sequential outputs $\mathbf{y}=[y_{s}, y_{1},...,y_{T-1}]$ is added to Equation (\ref{eq_3}).
\begin{equation}
(h_{t}, c_{t})=\left\{\begin{array}{l}
LD-LSTM(g_{f}, h_{f}, y_{s}) \qquad \quad \ \  t=1\\
LD-LSTM(g_{t-1}, h_{t-1}, y_{t-1}) \quad t>1
\end{array}\right.
\end{equation}

Next, the Multi-Head General Attention mechanism is performed to generate the glimpse vector related to the current hidden state $h_{t}$.
\begin{equation}
g_{t} = MultiHead(h_{t}, \mathbf{v})
\end{equation}

Finally, the predicted character is calculated by:
\begin{equation}
y_{t} = \phi(h_{t},g_{t}) = argmax(softmax(W_{o}[h_{t};g_{t}] + b_{o}))
\end{equation}

The hidden state $h_{t}$ and current glimpse vector $g_{t}$ are concatenated and passed through the linear transformation to make the final prediction to 63 classes, including 10 digits, 52 case sensitive letters and the 'EOS' token. Then the hidden state, cell state and glimpse vector are recovered and passed to the next step. The decoder works iteratively until the 'EOS' token is predicted or the maximum time step T is arrived.

Attention based sequence-to-sequence models\cite{shi2018aster,litman2020scatter,liao2019mask} generally use the decoding workflow based on \cite{bahdanau2014neural}, which is $h_{t-1}\rightarrow \mathbf{a_{t}} \rightarrow g_{t}\rightarrow h_{t}$, where the glimpse vector $g_{t}$ is related to the last time step and cannot be directly used for the current prediction. On the other hand, the workflow of RCEED is more intuitive, which is $(h_{t-1},g_{t-1})\rightarrow h_{t}\rightarrow \mathbf{a_{t}} \rightarrow g_{t}$. The decoding workflow of SAR\cite{li2019show} is comparable to us. Nevertheless, the first round RNN iteration of SAR does not make predictions, and its glimpse vector is not utilized in the RNN calculation. In our design, $\mathbf{g}$ as representation of a certain part of the encoded feature directly participates in the recurrent iteration of $\mathbf{h}$ and the character prediction. Therefore, a content enriched expression is obtained and the correlation between the visual feature and the character sequence is enhanced. 
\section{Experiments}
\subsection{Datasets}
We train our RCEED on part of the two synthetic datasets: MJSynth and SynthText, and the training sets of the four scene text datasets, i.e., IIIT5K, ICDAR2013, SVT, ICDAR2015. Evaluation results are based on the six standard benchmarks, including three regular datasets and three irregular datasets.

\textbf{MJSynth}(MJ)\cite{jaderberg2014synthetic} is also named Synth90k. MJ has 9 million synthetic images of english words, each of them is annotated with a word-level ground truth. We use 4.8 million images from MJ for training.

\textbf{SynthText}(ST)\cite{gupta2016synthetic} is another synthetic dataset which is widely used in text detection and recognition. We crop the word patches from the background images and random select 4.8 million out of 7 million word images for training.

\textbf{IIIT5K}\cite{mishra2012scene} is a natural scene text dataset cropped from the Google search images. It is divided into 2000 training and 3000 testing images.

\textbf{SVT}\cite{wang2011end} consists of 257 training and 647 testing data cropped from the Google Street View images. SVT is challenging for the blur or noise interference.

\textbf{ICDAR2013}\cite{karatzas2013icdar} contains 848 images for training and 1095 images for evaluation. We filter out the images that only contain the non-alphanumeric characters, resulting in a test set with 1078 images. 

\textbf{ICDAR2015}\cite{karatzas2015icdar} has 4468 word patches for training and 2077 for testing. IC15 is a benchmark dataset for irregular scene text recognition. The images are captured from the Google Glasses under arbitrary angles. Therefore, most words are irregular and have a changeable perspective.

\textbf{SVT-Perspective}(SVTP)\cite{phan2013recognizing} has 645 cropped scent text images for testing. It is also from the Google Street View but has more perspective words than SVT.

\textbf{CUTE80} (CUTE)\cite{risnumawan2014robust} consists of 288 word patches. CUTE is widely used for STR model evaluation for various curved text images.

\subsection{Implementation Details}
We build our model on the Tensorflow framework. All experiments are conducted on 2 NVIIDA Titan X GPUs with 12GB memory. The training dataset consists of 4.8 million from MJ, 4.8 million from ST, and 50k from the mixed training sets of IIIT5K, IC13, IC15 and SVT. Data augmentation methods such as perspective distortion and color transformation are applied to the mixed real datasets. The batch size is set to 42, including 20 from the MJ, 20 from the ST and 2 from the mixed dataset. The training process only needs 2.2 epochs in total with word-level annotation. Adam is utilized for optimization, with the addition of cross-entropy loss and l2 regularization loss(with the coefficient of 1e-4) as the objective function. Learning rate is set to 1e-4 per GPU, and reduced to 1e-5 in the last 0.2 epoch.

The input size of our model is $48\times160$. Height is fixed, if the image width is shorter than target width after resize with the original scale, we apply the padding operation, else we change the scale and simply reshape the image to the target size. The LD-BiLSTM in the encoder and LD-LSTM in the decoder are both single-layer, with the dropout rate of 0.1 and 0.5 respectively. The dimensions of the visual feature/context feature/decoded hidden state/word embedding matrix are set to 512/512/512/256. The maximum time step T is 27.

At the inference stage, no lexicon is used. For the sake of efficiency, we simply rotate 90 degree clockwise for the images of which the height is larger than 2 times of the width, instead of applying two directions rotation as previous studies\cite{shi2018aster,litman2020scatter,li2019show,lu2019master,yue2020robustscanner}. Besides, unlike\cite{shi2018aster,li2019show}, we do not use the beam search method to improve the performance. The decoder works in the forward direction, rather than the bidirectional strategy which is adopted in\cite{shi2018aster,yang2020holistic}.

\begin{table}[]
\vspace{-0.3cm} 
\caption{Recognition accuracy of our method and other sate-of-the-art methods on the six public datasets. All the results listed are lexicon free. The approaches which need character-level annotations are marked *. In each column, the best performance is shown in $\mathbf{bold}$ font, and the second best result is  \underline{underlined}. Our model achieves the best accuracy in regular dataset SVT and the most challenging irregular dataset IC15, the second highest accuracy in regular dataset IC13 and curve dataset CUTE.}
\label{tab2}
\centering
\begin{tabular}{|c|c|c|c|c|c|c|}
\hline
\multirow{2}{*}{\textbf{Method}} & \multicolumn{3}{c|}{\textbf{Regular Text}}     & \multicolumn{3}{c|}{\textbf{Irregular Text}}  \\ \cline{2-7} 
                                 & \textbf{IIIT5K} & \ \textbf{SVT} \  & \ \textbf{IC13} \  & \ \textbf{IC15} \ &  \ \textbf{SVTP} \ &  \ \textbf{CUTE} \ \\ \hline
CRNN(2015)\cite{shi2016end}                     &   81.2           &  82.7          &  89.6         & -             & -             & -             \\
$R^2AM$(2016)\cite{lee2016recursive}         & 78.4            & 80.7         & 90.0          &  -          & -         &  -          \\
RARE(2016)\cite{shi2016robust}         & 81.9            & 81.9         & 88.6          &  -          & -         &  -          \\
FAN(2017)\cite{cheng2017focusing}*                    & 87.4            & 85.9         & 93.3          &  70.6          & -         &  -          \\
NRTR(2017)\cite{sheng2019nrtr}                     & 86.5            & 88.3         &  \underline{94.7}          & -             & -             & -             \\
EP(2018)\cite{bai2018edit}*         & 88.3            & 87.5         & 94.4          &  73.9          & -         &  -          \\
ESIR(2018)\cite{zhan2019esir}                    & 93.3            & 90.2         & 91.3          & 76.9          & 79.6          & 83.3          \\
Char-Net(2018)\cite{liu2018char}*                 & 92.0            & 85.5         & 91.1          & 74.2          & 78.9          & -             \\
Mask TextSpotter(2018)\cite{liao2019mask}*         & \underline{95.3}           & $\mathbf{91.8}$         & $\mathbf{95.3}$          & 78.2          & 83.6          & 88.5          \\
ASTER(2018)\cite{shi2018aster}                     & 93.4            & 89.5         & 91.8          & 76.1          & 78.5          & 79.5          \\
ZOU, L et al.(2019)\cite{zuo2019natural}        & 85.4            & 84.5         & 91.0          &  -          & -         &  -          \\
MORAN(2019)\cite{luo2019moran}                   & 91.2            & 88.3         & 92.4          &  68.8          & 76.1         &  77.4          \\
MASTER(2019)\cite{lu2019master}                   & 95.0            & 90.6         & $\mathbf{95.3}$         & 79.4          & \underline{84.5}          & 87.5          \\
SCATTER(2020)\cite{litman2020scatter}                & 92.9            & 89.2         & 93.8          & \underline{81.8}          & \underline{84.5}          & 85.1          \\
Yang, L et al.(2020)\cite{yang2020holistic}            & 94.7            & 88.9         & 93.2          & 74.0          & 80.9          & 85.4          \\ \hline
$\mathbf{SAR(2019)}$\cite{li2019show}      & 95.0            & \underline{91.2}         & 94.0          & 78.8          & $\mathbf{86.4}$          & 89.6          \\
$\mathbf{SAR \ 3\times20}$    &94.0      &90.6      &93.1      &76.2     &83.7      &87.5     \\
$\mathbf{RobustScanner(2020)}$\cite{yue2020robustscanner}         & $\mathbf{95.4}$            & 89.3         & 94.1          & 79.2          & 82.9          & $\mathbf{92.4}$          \\ \hline
$\mathbf{RCEED(Ours)}$                       & 94.9                & $\mathbf{91.8}$             & \underline{94.7}              &  $\mathbf{82.2}$             &  83.6             &  \underline{91.7}             \\ \hline
\end{tabular}
\end{table}
\vspace{-0.5cm}

\subsection{Experimental Results}
In this section, we test the model performance on regular and irregular STR public datasets, and make comparison with the SOTA methods. As shown in Table \ref{tab2}, RCEED achieves the best performance in regular dataset SVT and irregular dataset IC15, and the second highest accuracy in regular dataset IC13 and curve dataset CUTE. Compared with competitors SAR\cite{li2019show} and RobustScanner\cite{yue2020robustscanner} which are baseline and up-to-date methods targeting at irregular scene text recognition and trained with synth\&real datasets, RCEED performs accuracy increases of 3pp to 3.4pp in the most challenging irregular dataset IC15 and outperforms them by 4 out of 6 benchmarks, while the size of the encoded feature map is only 1/4 of theirs. Compared with SAR modeled with the same-size encoded feature map(SAR 3$\times$20), our model outperforms it in 5 benchmarks with 6pp ahead in maximum. Experimental results demonstrate that RCEED has got rid of the performance bottleneck caused by small-size feature map, and is competitive in both the regular and the irregular datasets.

\begin{figure}
\vspace{-0.3cm} 
\setlength{\belowcaptionskip}{-0.5cm} 
\centering
\includegraphics[width=\textwidth]{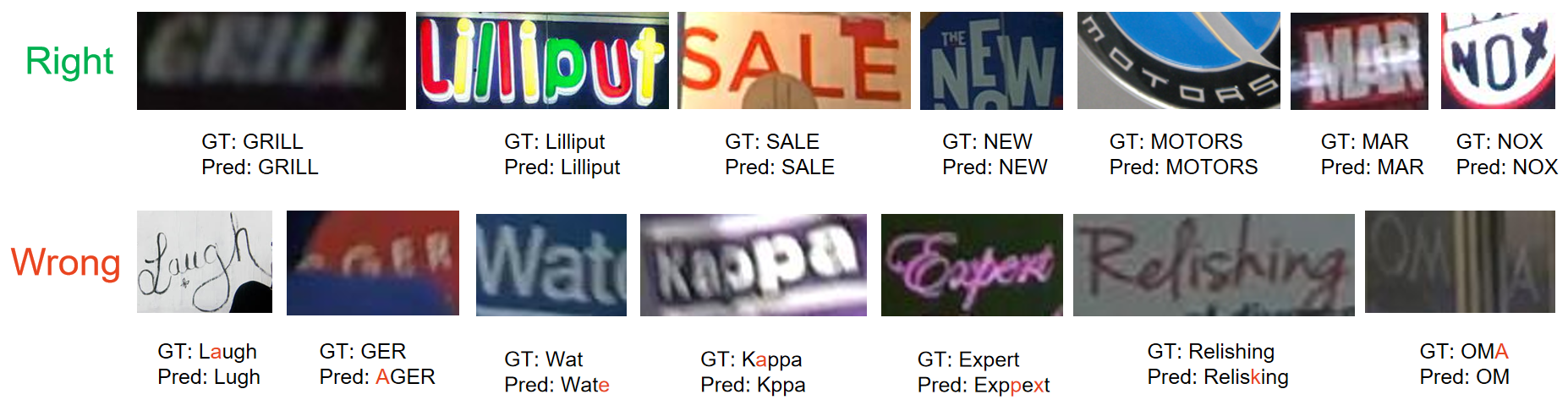}
\caption{Success and failure cases of RCEED. GT refers to the ground truth, Pred refers to the predicted results. } 
\label{Fig.5}
\end{figure}

Figure \ref{Fig.5} shows the success and failure cases of RCEED. Success cases in the first row demonstrate that the proposed model has a robust ability to deal with blur, variation, occlusion, distortion, uneven light and other difficult situations. However, when the problems become serious, failure cases such as over-prediction and less-prediction appear. It can be seen that image quality still has a significant impact on the recognition results.

\subsection{Ablation Studies}
In this section, we conduct a series of comparison experiments to analyse the effectiveness of the key contributions in RCEED. Evaluation results of 8 conditions are listed in Table \ref{tab3}.

\noindent\textbf{Impact of comprehensive feature} \quad 
We evaluate the performance when visual feature and context feature are used alone as the input of the decoder. As shown in Row 2 and 3, single visual and single context features both have lower accuracy of up to -1.6pp on the regular dataset and -2.8pp on the irregular dataset compared with the comprehensive feature, indicating that the feature fusion operation plays an important role in improving model performance.   

\noindent \textbf{Impact of Layernorm-Dropout LSTM Cell} \quad
Row 1 reveals the model performance when the LD-LSTM design in the encoder and decoder are both replaced with normal LSTM cells. Compared with Row 8, the model implemented with normal LSTM cells shows an accuracy drop of -1.3pp on the largest irregular dataset IC15. While in the largest regular dataset IIIT5K, the accuracy is 0.6pp higher and equals to the best performance\cite{yue2020robustscanner}. The dropout design in the LD-LSTM improves model's ability in dealing with changeable characters, but is not compatible with regular texts with stable features. Overall, LD-LSTM cell still outperforms the normal one in most benchmarks. 

\noindent \textbf{Impact of guided initialization of decoder} \quad
Row 4 lists the results when the decoder is initialized with zero vectors instead of the holistic feature and global glimpse vector from the encoder. Accuracy shows a drop up to -1.6pp in the five out of six public datasets, which demonstrates that the global information from the encoder is helpful for improving the decoding accuracy. 

\noindent \textbf{Impact of glimpse vector for prediction} \quad
We build a comparison model which does not use glimpse vector for prediction. As shown in Row 5, the performance degradation is more serious in irregular datasets(up to -1.8pp), proving that visual information has a greater impact on recognizing complex texts.

\noindent\textbf{Impact of heads number} \quad
We compare the evaluation results when attention heads number m is set to 1, 4 and 8 in Row 6, 7, 8. The 1 head condition is equivalent to not using the multi-head design. We set m=8 in our model for the superior performance on irregular datasets.

\begin{table}[]
\vspace{-0.3cm} 
\caption{Ablation studies by changing model structures and hyper-parameters.  LD refers to models w/o the Layernorm-Dropout method in the RNN layers. VF/CF denote whether the Visual Feature and Context Feature are involved in the encoded feature map. GI/GP analyze the impact of the Guided Initialization of decoder and the Glimpse vector for Prediction. Heads column presents the conditions with different number of heads. All the comparison models are trained from scratch with word-level annotations.}
\label{tab3}
\centering
\begin{tabular}{|c|c|c|c|c|c|c|c|c|c|c|c|c|}
\hline
\textbf{Cond} & \textbf{LD} & \textbf{VF} & \textbf{CF} & \textbf{GI} & \textbf{GP} & \textbf{Heads} & \textbf{IIIT5K} & \textbf{SVT} & \textbf{IC13} & \textbf{IC15} & \textbf{SVTP} & \textbf{CUTE} \\ \hline
1             & $\times$                & $\checkmark$    & $\checkmark$       & $\checkmark$           & $\checkmark$          & 8              & 95.4            & 91.5         & 94.7          & 80.9          & 84.9          & 89.9          \\
2             & $\checkmark$        & $\times$      & $\checkmark$   & $\checkmark$        & $\checkmark$          & 8              & 94.7            & 90.9         & 93.1        & 80.0            & 82.6          & 88.9          \\
3             & $\checkmark$        & $\checkmark$       & $\times$    & $\checkmark$       & $\checkmark$          & 8              & 94.9            & 91.6         & 93.7          & 80.4          & 82.8          & 90.3          \\
4       & $\checkmark$        & $\checkmark$        & $\checkmark$     & $\times$      &$\checkmark$           & 8              & 95.3            & 90.9         & 94.6          & 80.6          & 83.3          & 90.6          \\
5             & $\checkmark$        & $\checkmark$        & $\checkmark$     & $\checkmark$      & $\times$          & 8              & 94.9            & 92.6         & 94.3          & 80.9          & 82.9          & 89.9          \\
6             & $\checkmark$      & $\checkmark$       & $\checkmark$    & $\checkmark$       & $\checkmark$          & 1              &94.3                 &91.6              & 94.3              &  80.8             & 83.6              & 91.7              \\
7     & $\checkmark$      & $\checkmark$   & $\checkmark$     & $\checkmark$      & $\checkmark$          & 4              & 95.0              & 92.4         & 94.1          & 81.1          & 83.7          & 89.5          \\  \hline
8    & $\checkmark$      & $\checkmark$      & $\checkmark$  & $\checkmark$   & $\checkmark$          & 8             & 94.9            & 91.8       & 94.7          & 82.2         & 83.6         & 91.7        \\ \hline
\end{tabular}
\end{table}
\vspace{-0.5cm}

\section{Conclusion}
In this work, we propose a representation and correlation enhanced encoder-decoder framework for scene text recognition. The encoder enhances model's representation ability through the aligning and fusing operation between the local visual feature and the global context feature. The decoder strengthens the correlation between the encoded comprehensive feature and the decoded character sequence through the guided initialization and the efficient workflow. Essential components including the Multi-Head General Attention mechanism and the LD-LSTM cell are designed to reduce feature deficiency and improve the generalization towards changeable texts. The model breaks the constraint of feature map size and has a superior performance on public benchmarks. In future research, we will develop an end-to-end integrated detection and recognition model for the text spotting task and develop advanced applications of the visual-semantic interaction.

\section{Acknowledgments}
This work is supported by National Natural Science Foundation of China (61976214, 61721004, 61633021), and Science and Technology Project of SGCC Research on feature recognition and prediction of typical ice and wind disaster for transmission lines based on small sample machine learning method.
\vspace{1cm}

\bibliographystyle{unsrt}
\bibliography{new}

\end{document}